\documentclass[letterpaper, 10 pt, conference]{ieeeconf}  

\IEEEoverridecommandlockouts                              

\usepackage{graphics} 
\usepackage{graphicx}
\usepackage{mathptmx} 
\usepackage{times} 
\usepackage{amsmath} 
\usepackage{amssymb}  
\usepackage{color}
\usepackage{algorithm}%
\usepackage{algorithmicx}%
\usepackage{algpseudocode}%
\usepackage{comment}
\usepackage{subfig}
\usepackage{float}
\usepackage[noadjust]{cite}
\usepackage[normalem]{ulem}
\usepackage[T1]{fontenc}
\usepackage[bottom=72pt, top=54pt, left=45pt, right=45 pt ]{geometry} 
\definecolor{mygray}{rgb}{0.5, 0.5, 0.5}
\definecolor{orange}{rgb}{1.0, 0.5, 0.2}

\title{\LARGE \bf
Developmental Scaffolding with Large Language Models 
}

\author{
    Batuhan Celik$^1$, Alper Ahmetoglu$^1$, Emre Ugur$^1$, Erhan Oztop$^{2,3}$
    \thanks{ $^{1}$ Computer Engineering Dept., Bogazici University, Istanbul, Turkey.}
    \thanks{ $^{2}$ Computer Science Dept., Ozyegin University, Istanbul, Turkey.}
    \thanks{ $^{3}$ Symbiotic Intelligent Systems Research Center, Institute for Open and Transdisciplinary Research Initiatives, Osaka University, Japan.} 
    }
    
\usepackage{hyperref}
\begin{document}

\maketitle
\thispagestyle{empty}
\pagestyle{empty}

\begin{abstract}
Exploration and self-observation are key mechanisms of infant sensorimotor development. These processes are further guided by parental scaffolding {to accelerate} skill and knowledge acquisition. In developmental robotics, this approach has been adopted often by having a human acting as the source of scaffolding. In this study, we investigate whether Large Language Models (LLMs) can act as a scaffolding agent for a robotic system that aims to learn to predict the effects of its actions. To this end, an object manipulation setup is considered where one object can be picked and placed on top of or in the vicinity of another object. The adopted LLM is asked to guide the action selection process through algorithmically generated state descriptions and action selection alternatives in natural language. The simulation experiments that include cubes in this setup show that LLM-guided (GPT3.5-guided) learning yields significantly faster discovery of novel structures compared to random exploration.
However, we observed that GPT3.5 fails to effectively guide the robot in generating structures with different affordances such as cubes and spheres.  Overall, we conclude that even without fine-tuning, LLMs may serve as a moderate scaffolding agent for improving robot learning, however, they still lack affordance understanding which limits the applicability of the current LLMs in robotic scaffolding tasks.
\end{abstract}

\section{INTRODUCTION}
\label{intro}
In robotics, random exploration is a common mechanism used in learning from reinforcement signals \cite{mnih2013playing, Ladosz_2022}, navigation \cite{6696676}, and manipulation and planning \cite{ahmetoglu2022learning}. However, in complex environments where action-to-effect mapping is non-linear, stochastic, and/or redundant, hard-to-reach states may not be experienced \cite{ten_oudeyer_moulin-frier_2021}. For example, in our previous work on effect prediction and planning, as the number of objects increases and the action set is extended,
creating composite structures such as bridges, T-shaped and U-shaped structures becomes possible \cite{ahmetoglu2022learning}.
However, constructing such complex structures requires performing a series of \emph{correct} actions, which {becomes less and less likely} to experience with random exploration as the complexity of the setup increases.

One approach regarding exploring complex environments is extracting the required knowledge from humans with methods like imitation \cite{SCHAAL1999233,pmlr-v24-vlachos12a} and parental scaffolding \cite{ugur_nagai_celikkanat_oztop_2015,ugur2011}. While data collection from humans can be assumed as the gold standard for imitation learning and parental scaffolding, it may become a labor-intensive process for complex learning tasks. 

With the introduction of large language models (LLMs), it became possible to generate human-like natural text and code-like structured text for various tasks \cite{openai2023gpt4, chowdhery2022palm, radford2021learning}. This recent success is fueled by the internet-scale datasets \cite{brown2020language,touvron2023llama} together with efficient deep learning architectures such as Transformers \cite{vaswani2017attention} that can squeeze out information from a vast amount of data. Although LLMs can be considered as a huge smart look-up table (see \cite{weng2023deep} for a critique), their applications in various fields\cite{törnberg2023chatgpt4, driess2023palm,zhang2023complete} has shown that they can be successfully applied to many practical tasks.

The sequence generation capabilities of LLMs are shown to be useful for robotic applications as well\cite{vemprala2023chatgpt, kira2022llmroboticspaperslist}. Additionally, being trained on internet-scale data, they can be utilized as knowledge bases \cite{ren2022leveraging, shridhar2021cliport}. By utilizing these two features, previous works \cite{saycan2022arxiv} \cite{driess2023palm} provide methods to utilize PALM \cite{chowdhery2022palm} as a robotic controller. Similarly, \cite{ren2022leveraging} and \cite{cui2022foundation} utilized CLIP \cite{radford2021learning} as a knowledge base for zero-shot learning. However, grounded knowledge is scarce on the internet, therefore, LLMs lack real world experience \cite{saycan2022arxiv, driess2023palm}. This poses a problem in LLM-based robot applications since inferences with incorrect grounding are usually unusable in real-life scenarios. Huang et al. \cite{huang2022language} demonstrate LLMs shortcomings in executable plan generation.

Conditioning the LLMs for the agent's embodiment is required to perform grounded inferences in robotic settings\cite{mees2023grounding}. Previous work suggests different strategies to perform \emph{descriptive embodiment alignment} such as bottom-up skill learning \cite{vemprala2023chatgpt,wang2023voyager}, prompt engineering \cite{wei2023chainofthought,singh2022progprompt}, and the use of multiple modalities \cite{driess2023palm,jin2023alphablock,shridhar2021cliport}. Additionally, the response selection \cite{kirk2023improving,saycan2022arxiv} can be used to eliminate non-aligned responses {and finetuning can increase grounded reasoning accuracy\cite{10.3389/frobt.2023.1221739}}. Another method for utilizing the LLMs' knowledge about the real world is LLM scaffolded exploration \cite{tam2023semantic,jin2023alphablock,lykov2023llmbrain}. As LLMs provide a large knowledge base about the real world, they can possibly lead the exploration process in developmental/continual learning settings. Additional networks can be trained with sensorimotor data collected from complex {exploration sessions} with the help of LLMs. These networks are expected to make grounded inferences since they are trained {using grounded knowledge} while being exposed to hard-to-reach states. Works \cite{lykov2023llmbrain} and \cite{jin2023alphablock} explore such processes where the data collection was controlled by GPT3 {\cite{brown2020language}} or GPT4 {\cite{openai2023gpt4}}. The collected data was used to fine-tune smaller multi-modal LLMs. Similarly, \cite{tam2023semantic} proposed a method that uses different LLMs to guide exploration during reinforcement learning while 
\cite{xiao2023robotic} utilizes LLMs to augment robot-language data.


In this work, we {demonstrate that} an LLM (GPT3.5) {can be used as} a parental scaffolding agent for a simulated robot that tries to discover the effects of its actions. Additionally, we devise a new token-efficient prompting strategy to allow the LLM to lead the exploration process {without descriptive embodiment alignment}. For requesting guidance for robot exploration, GPT3.5 is prompted to select an action that would result in an `interesting' or `novel' outcome and to produce the output in a fixed format that can be easily parsed. We tested our scaffolding agent against a random exploration baseline in tabletop environments with different {complexities} and observed that GPT3.5 can actively lead the exploration to hard-to-reach states. Additionally, we showed the effects of different prompt structures and tested GPT3.5's capability of making grounded inferences when different types of objects with different affordances \cite{cakmak,jamone} are introduced.

\section{Method}
\begin{figure}[htbp]
    \centering
    \subfloat[Step 1]{\includegraphics[width=0.235\textwidth, ]{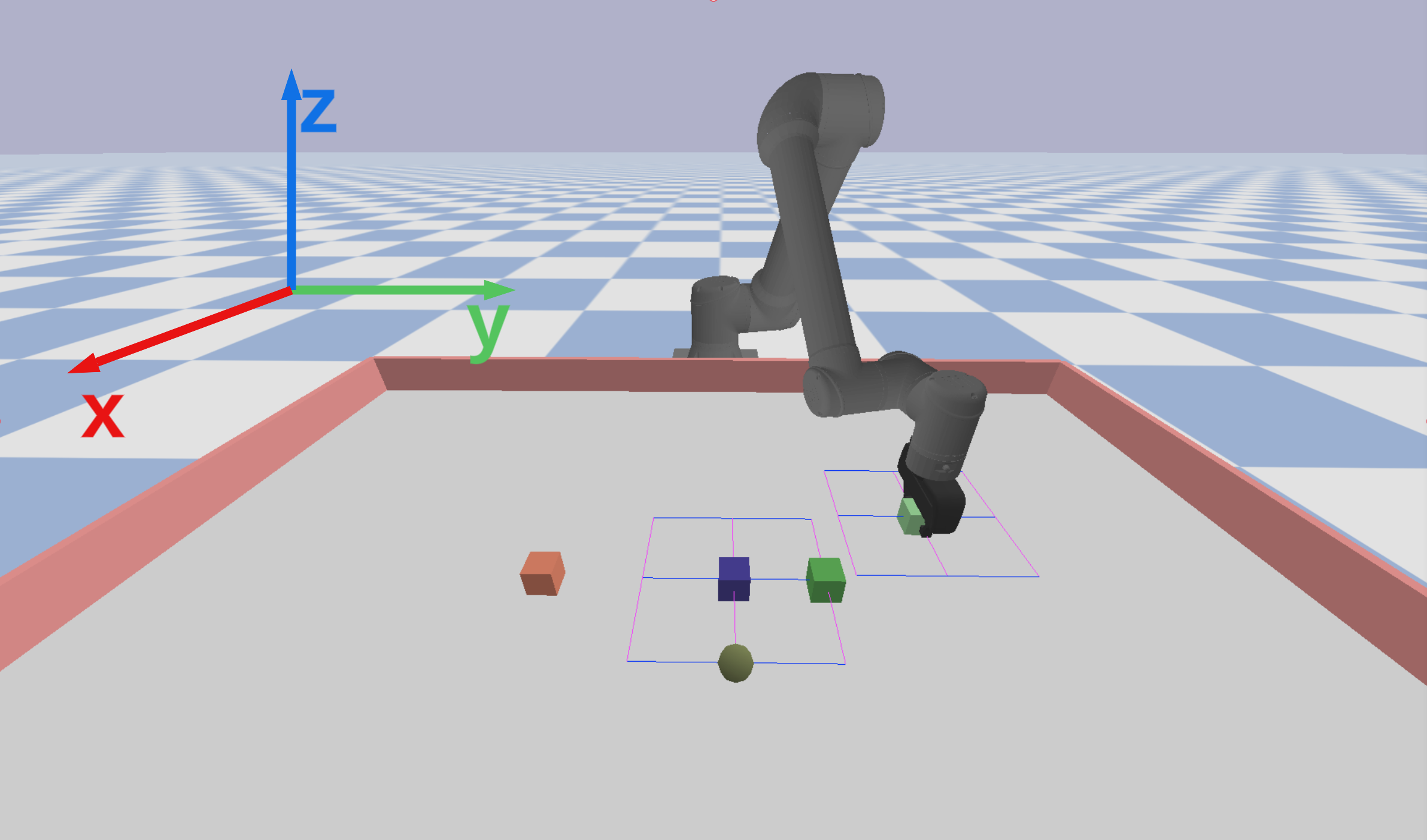}
    \label{fig:test-overall}}
    \hfill
    \subfloat[Step 2]{\includegraphics[width=0.235\textwidth, ]{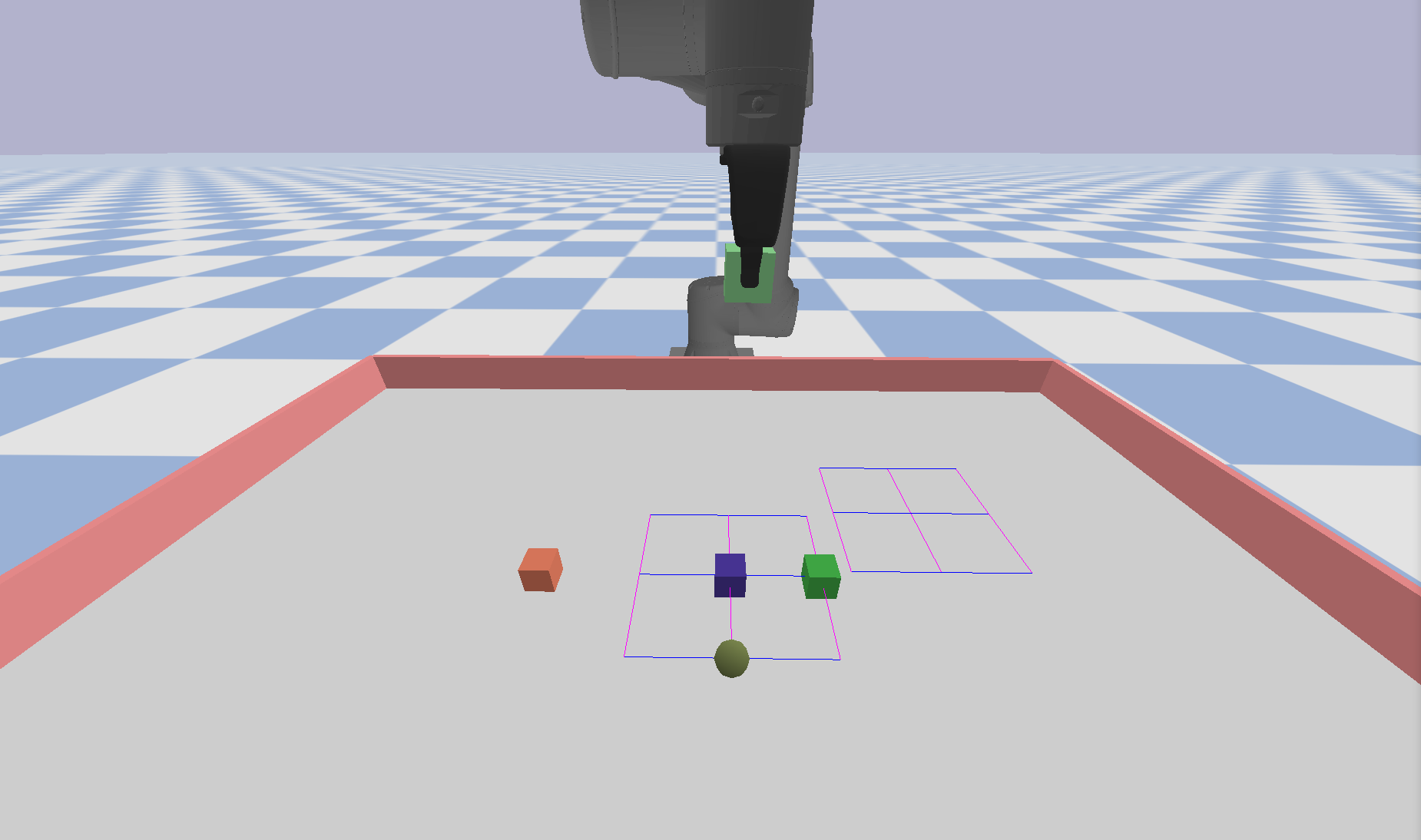}
    \label{fig:train-overall}}
    \vfill
    \subfloat[Step 3]{\includegraphics[width=0.235\textwidth, ]{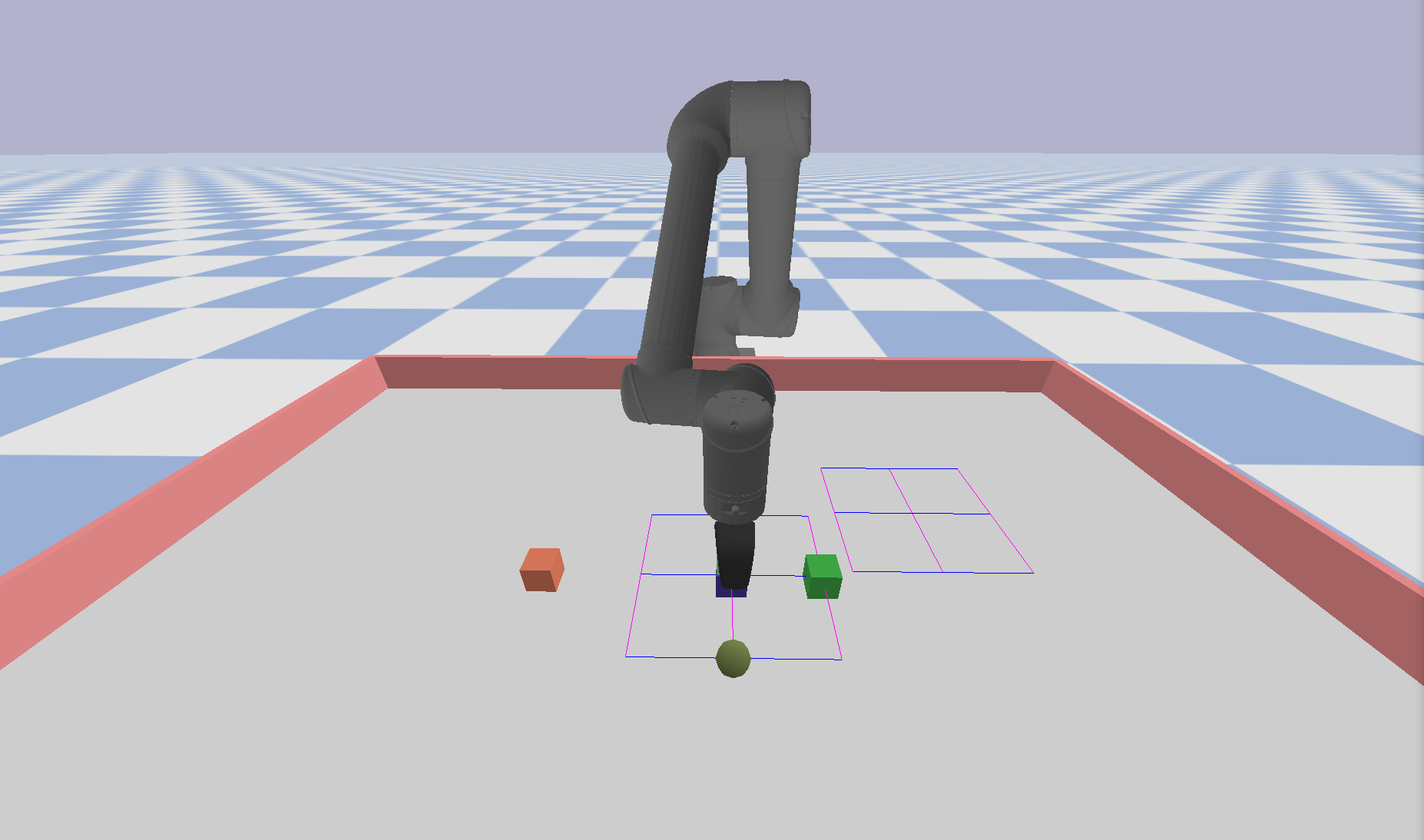}
    \label{fig:test-overall}}
    \hfill
    \subfloat[Step 4]{\includegraphics[width=0.235\textwidth, ]{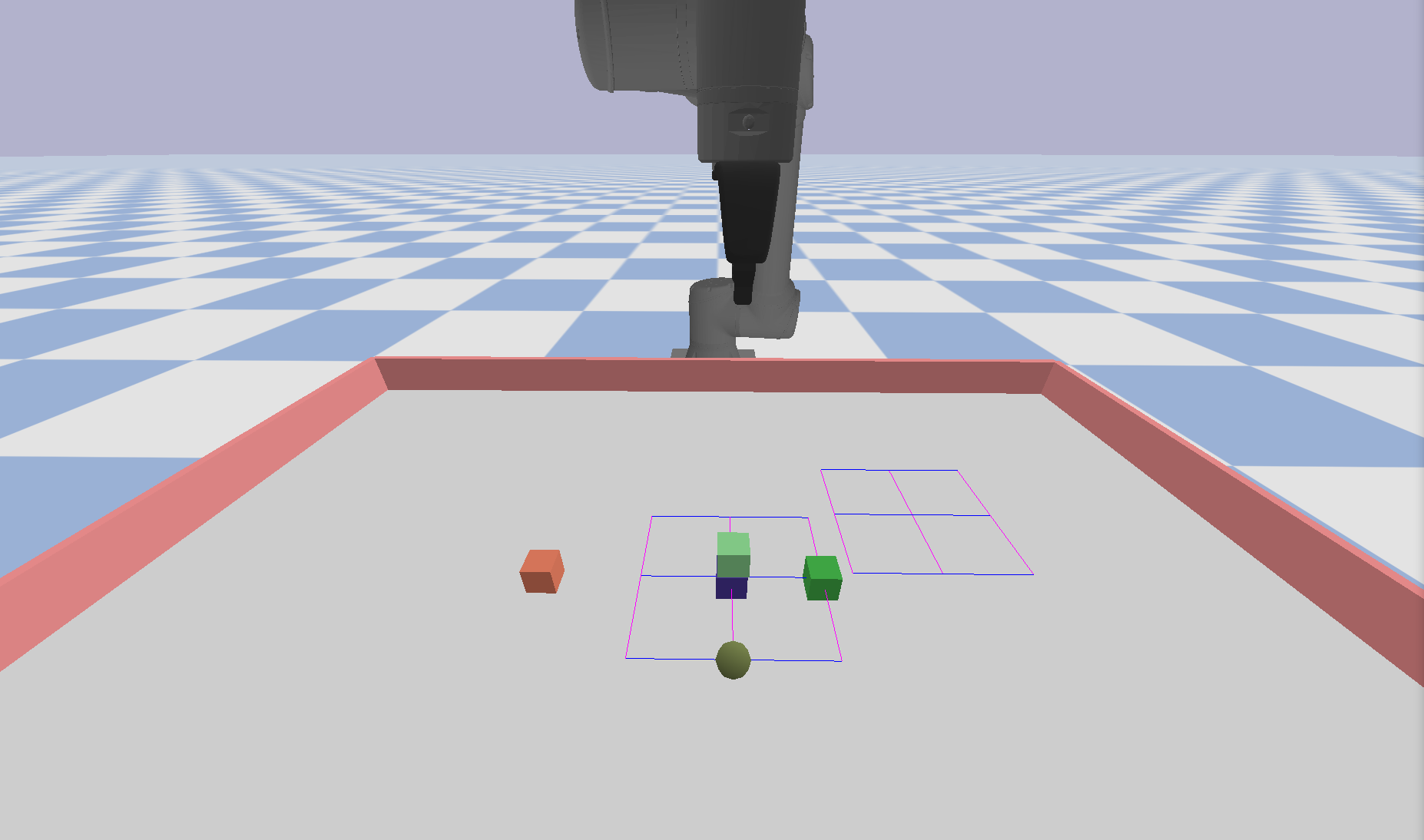}
    \label{fig:train-overall}}

    \caption{Action execution steps are shown in the subfigures. Grids around objects indicate possible grasp and placement locations. In our experiments, only the center location is used for grasping. As for placement locations, the next, front, and center locations are used. Prior to the execution of this action, the sphere is placed in front of the purple cube, and the dark green cube is placed next to the purple cube. }
    \label{fig:action}
\end{figure}

Inspired by skill acquisition via parental scaffolding during infant development, we propose a robotic scaffolding method that regards LLM as a scaffolding agent akin to a caregiver that guides the learning exploration of a robotic agent. Note that infant scaffolding may take the form of kinesthetic guidance, attention directing, or choosing the right choice for the infant for a given task. In this work, we focus on the latter type of scaffolding and propose a method to exploit the high-level knowledge contained in LLMs to provide automated scaffolding to accelerate learning.

Even though foundation models can not be used to perform accurate grounded inferences directly \cite{saycan2022arxiv, driess2023palm, huang2022language}, we hold that the knowledge harvested from the internet may allow LLMs to scaffold robot exploration. {As such, unlike other applications of LLMs, we do not apply fine-tuning.} For example, in the case of a tabletop manipulation environment, the ChatGPT recognizes balancing a cube on top of a sphere as a challenging task {in its foundation model form}. Thus, our expectation is that GPT3.5 can distinguish among alternative action choices and make suggestions for better exploration {without further conditioning on the real world}. 

As for the knowledge extraction method, the LLM is expected to steer the exploration towards rare but desired task configurations by selecting actions that have more exploration potential.
{The collected sensorimotor data thus is expected to contain hard-to-reach states that are important for learning action-to-effect mapping to allow real world reasoning.} 
Finally, the result of the learning with the collected data is expected to create a system that can make grounded inferences.


To show the validity of LLM-based scaffolding, a simulated robot is equipped with an action repertoire in a tabletop object manipulation environment where it experiments with its action capabilities and collects data for learning the effects of its actions. In this environment, hard-to-reach task configurations are complex structures that can be built from basic objects such as cubes and spheres. In this setup, experiences are collected {in several environment configurations} using different exploration policies including a random baseline to assess the efficiency of the proposed method. The visitation frequency to hard-to-reach states such as tall towers is used as a measure of exploration efficacy gained by LLM-based scaffolding.



\subsection{Experiment Environment}
A simulated robot, UR10 (Universal Robots), and a tabletop environment are used to capture the behavior of an infant exploring its environment. The tabletop environment serves as a confined setup where the task complexity can be tuned by changing the number of available actions and objects. The robot is equipped  with high-level pick-and-place actions to interact with the objects on the table. The 
 robot and its interactions with the environment are simulated with the PyBullet simulation library. The robot interacts with the objects in `{sessions}' which consists of 10 interactions.
 Each {session} begins with various numbers of {cubes and spheres randomly initialized on the table. The cubes have an edge size of 2.5 cm while the spheres have a diameter of 2.5 cm}. Then, ten pick-and-place actions are executed within each session. At each interaction step, the robot selects a source and a target object to interact with and executes a pick-and-place operation by first grasping the source object and then placing it on top of or next to the target object. 

In brief, the actions available to the robot are characterized by 3 integer tuples: $(o_s, o_t, p)\: s.t. \: o_s \neq o_t$ where $o_s$ indicates the grasped object {id}, $o_t$ indicates the target object {id} and $p$ indicates the discrete relative positions. Three possible relative locations are allowed for the placement operation: the top ($p = 0$), next ($p = 1$), and front ($p = 2$) positions. The top position corresponds to direct stacking operation whereas the next and front positions correspond to a displacement of 15 cm in the positive y and x directions. The actions of the robot are generated  by first computing desired joint angles corresponding to desired robot locations by the inverse kinematics solver of PyBullet, and then running a joint tracking loop. 
An example execution is illustrated in Figure \ref{fig:action}. 

\subsection{LLM interfacing for scaffolding}
\label{sec:methods_llm}
Scaffolding for robot action exploration is tested in sessions of 10 object interactions where the simulated robot receives action suggestions from the LLM. To this  end, the current object configuration is described to the LLM in natural language using algorithmically generated prompts. Similarly possible action choices are algorithmically generated for LLM to choose from. 
\subsubsection{GPT system prompt definition} As our system is designed for the GPT3.5, system prompts are available to us in addition to the user prompts. A system prompt is a special type of prompt that is used for conditioning GPT3.5 to follow a specific behavior described in a natural language during the conversation {\cite{ouyang2022training}. Therefore, we used the system prompt to provide the task definition and the output format to GPT3.5. In order not to introduce any bias towards a specific selection,  the task definition simply consists of selecting an action with an interesting outcome in the given environment configuration. This short definition results in GPT3.5s' selecting actions by referring to the information it is trained on.
As for the output definition, the GPT3.5 is conditioned to provide the reasoning behind its selection before making a decision. Generating the reasoning before the decision results in less biased reasoning and increases the success rate of the GPT in robotics tasks \cite{vemprala2023chatgpt} as the generated tokens are conditioned on the previous tokens. 
The complete system prompt is the following:\\

\vspace{1mm}
\noindent\fbox{%
    \parbox{0.9\linewidth}{\small%
\texttt{[System]}: There are some objects on the table. Which manipulation alternative on them yields an interesting outcome? Choose one and explain.

Your output should be in the following format:

<reasoning> some sentences </reasoning>

Selected action is : <number of the selected action>
}}
\vspace{2mm}

\subsubsection{GPT User Prompt Definition}
The user prompt is conventionally used to query an answer from the GPT. In this study, we used it to inform GPT about the environment configuration. To do so, our system generates a prompt as follows:  $<S_i><H_{1,...,i-1}><A_{1, ...,k}>$, where $i$ indicates the number of previously selected interactions, $S_i$ is the description of the current configuration, $H_{1,...,i-1}$ is the session history, and $A_{1, ...,k}$ are possible actions to execute, all given in natural language. 
The configuration description contains a list of objects and the spatial relations between objects. Colors are used as unique object identifiers while describing the scene. 
The session history contains a list of previously executed actions in an orderly manner. Instead of conducting a dialogue between GPT3.5 and the robot during the whole interaction session, a new dialogue is initialized for each configuration, and session history is summarized in  $H_{1,...,i-1}$. 
Finally, a list of possible actions is provided to GPT3.5 to choose from. By providing a set of actions from the action repertoire, our system ensures that the selected action will be within the execution capability of the robot. 
An example prompt that describes the configuration given in step 4 of Figure \ref{fig:action} is as follows:
\\

\noindent\fbox{%
    \parbox{0.96\linewidth}{\small%
\texttt{[User]}: There is an orange cube, a green cube, a purple cube, a brown sphere, and a light green cube in the current scene.

the green cube is next to the purple cube.

the brown sphere is in front of the purple cube.

the light green cube is stacked on the purple cube.

Previously executed actions:

Put the brown sphere in front of the purple cube

Put the green cube next to the purple cube

Put the light green cube on top of the purple cube

\dots

Possible actions:

1 ) Put the green cube in front of the orange cube

2 ) Put the brown sphere next to the light green cube

3 ) Put the orange cube on top of the light green cube

\dots
    }
}
\\


\subsubsection{GPT Settings} For all of our experiments \verb|gpt-3.5-turbo| is used as the LLM. As for the API call parameters, the temperature is set to 0 to reduce the randomness of the answers. 

\section{Experiments and Results}
To assess the effectiveness of our method for scaffolding the exploration, a purely random action selection strategy is used as the baseline. If LLM indeed had some insight as to what actions yield better exploration of the configuration space, then it would discover complex structures, in particular towers, significantly faster than with a random exploration strategy. In addition to this experiment, several other experiments are conducted to shed light on the scaffolding capability provided by the LLM.

\subsection{Comparisons with the baseline exploration strategy}
\label{baseline}
For comparing the performance of our system against the baseline we first consider an environment with four cubes and two possible relative object locations (Experiment-1): on top of and next to. Each interaction session consists of ten interactions with the objects {led by} either by using LLM scaffolding or by using the random exploration strategy. The LLM is asked to pick the action that would lead to the most \emph{interesting} configuration at each interaction step. To account for the stochasticity in initial random object placement and multiple choice generation, the interaction session is repeated 40 times. Figure~\ref{4act_2loc} shows the results obtained in terms of the maximum tower height achieved. {Evidently,} LLM scaffolded exploration discovers the tallest possible tower much earlier than the random exploration case. 
   \begin{figure}[thpb]
      \centering
      \subfloat[Experiment 1: 4 cubes,  2 proximity locations]{\includegraphics[width=0.48\textwidth]{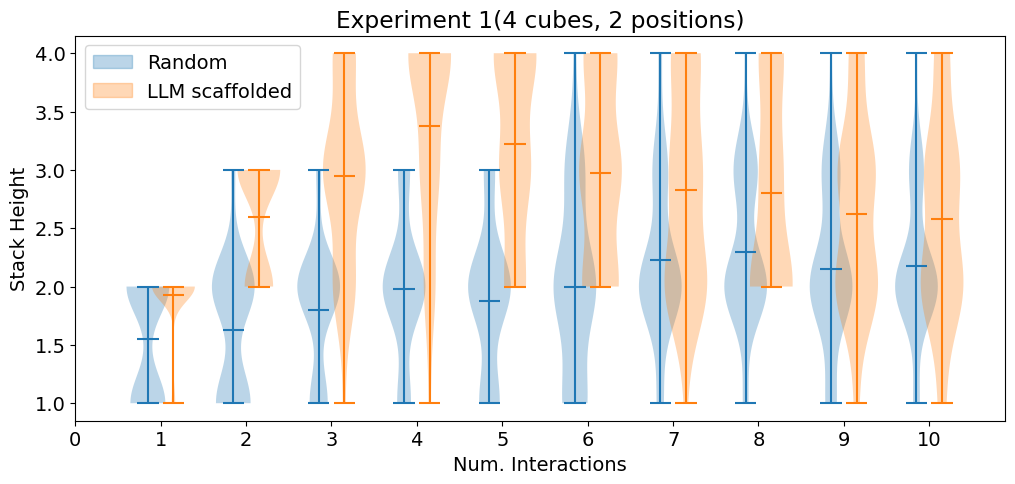}
      \label{4act_2loc}}\vfill
       \subfloat[Experiment 2: 5 cubes,  2 proximity locations]{\includegraphics[width=0.48\textwidth]{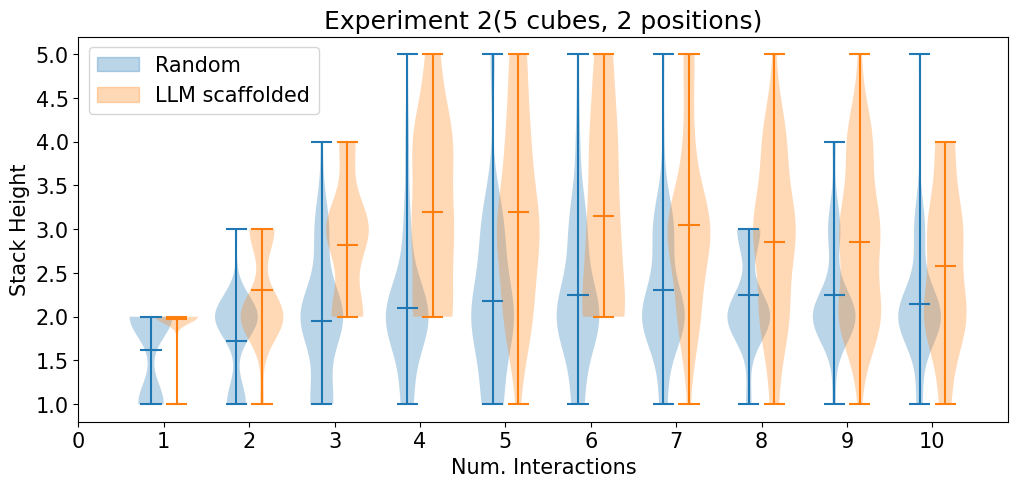}\label{5act_2loc}}\vfill
       \subfloat[Experiment 3: 5 cubes,  3 proximity locations]{\includegraphics[width=0.48\textwidth]{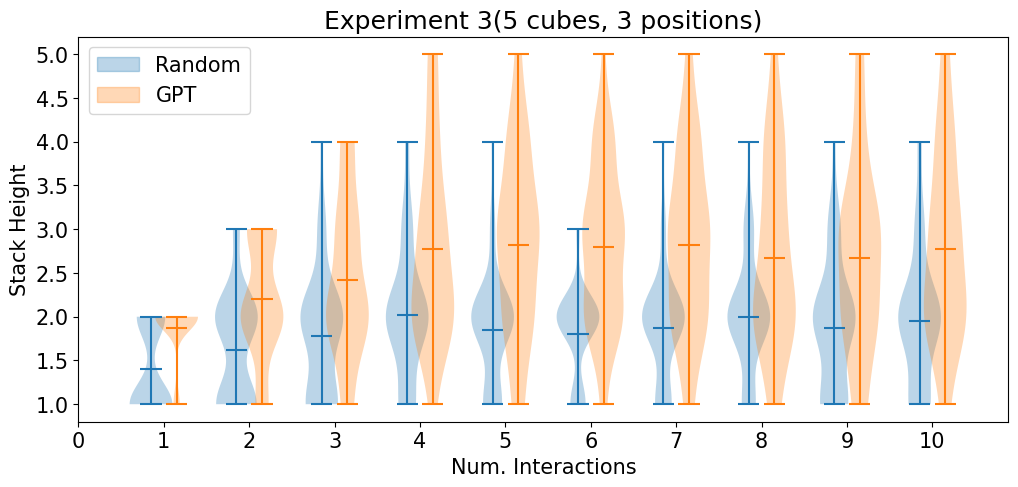}\label{5act_3loc}}\vfill
      \caption{Comparison of tower heights between random exploration and scaffolded exploration in different environment settings with incremental difficulty. The first setting contains 4 cubes and 2 positions, the second one introduces the fifth cube, and the last one introduces the third proximity location. }
      \label{2act}
   \end{figure}

To further test the generality of this result, in experiment two, the number of objects is increased to five. Figure~\ref{5act_2loc} shows the tower height distributions when an additional cube is introduced (Experiment-2). Also with this setup, LLM scaffolding based action selection obtains higher towers, in a shorter number of interactions. The random exploration generates the tallest possible tower in only one of the episodes. 


In Experiment-3, the number of placement actions that can be executed by the robot is increased from two to three (on top of, next to, and in front of) to demonstrate the effect of a larger action repertoire. Figure~\ref{5act_3loc} shows the comparison between LLM scaffolded exploration and random exploration in this configuration. As can be seen, the random exploration fails to reach a tower of height five. On the other hand, LLM scaffolding based exploration manages to create a tower of height of five in eight interaction sessions, indicating a significant contrast compared to the random exploration case.


As the complexity of the environment increases, the probability of visiting hard-to-reach states during random exploration decreases. This is exemplified by the failure of random exploration to achieve the tallest possible tower in the most complex environment setting (Experiment-3). 
On the other hand, LLM scaffolding facilitates the construction of tall towers at the initial phase of a session. However, this trend does not continue till the end of the session as can be seen in Figure~\ref{4act_2loc} and \ref{5act_2loc}. This is due to the action choices suggested by GPT3.5 in response to the query of choosing the action to create an interesting outcome and tall towers lose their novelty once encountered.


\subsection{Effects of different prompts}
\begin{figure}[thpb]
      \centering
      \includegraphics[width=0.48\textwidth]{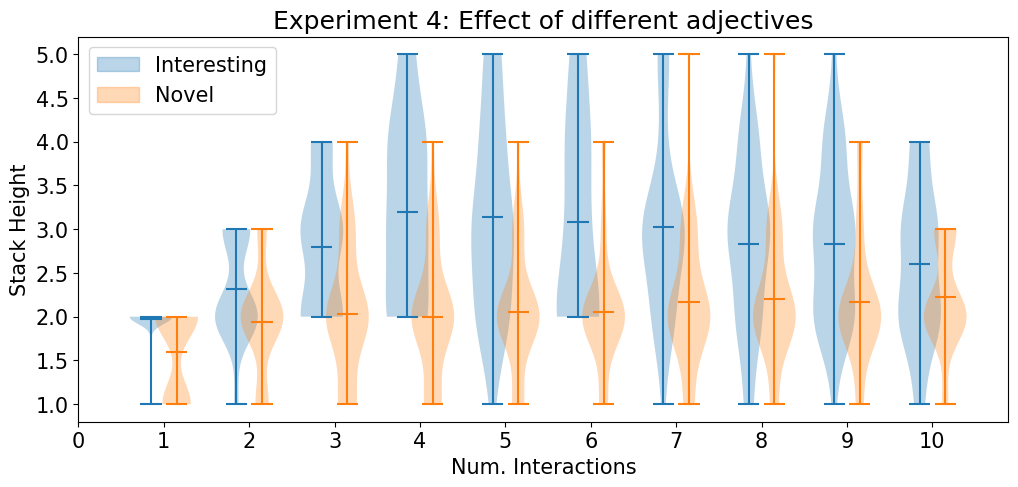}
      \caption{Effects of different adjectives on the tower heights in a 5 cubes 2 positions setting.}
      \label{int_vs_novel}
   \end{figure}

Experiments 1-3 probed GPT3.5's  ability to accelerate learning exploration when requested to select actions to create `interesting' outcomes. To assess the effect of the adjective `interesting', it is replaced with the word `novel' in the system prompt, and an LLM scaffolding experiment is conducted with a five-object, two-position configuration (Experiment 4). Figure~\ref{int_vs_novel} shows the average tower height attained for the `novel' adjective case superimposed with the `interesting' adjective case. Observation shows there is a notable difference in resulting tower heights. When GPT3.5 is conditioned to yield a `novel' outcome, it focuses on the history and actively avoids actions similar to the previously executed ones. Therefore, performing the stacking operation four consequent times is not likely in this setting as it requires performing consequent stacking operations, resulting in a low visit frequency to states with complex structures. These results show the importance of selecting the appropriate words when minimal prompting is used.

\subsection{Exploiting LLM's Affordance Knowledge}
   
Cubes are inherently  easily graspable and stackable, therefore, manipulating them does not require extensive knowledge regarding affordances.
In order to evaluate the GPT's performance when affordance is taken into account, we introduce a sphere into the environment alongside four cubes when all spatial positions are allowed. This setup allows us to examine how GPT3.5 adapts to the inclusion of a new object with distinct affordances. Similar to previous experiments, 40 sessions are collected with LLM scaffolded and random exploration policies. 

Figure \ref{sphere_rand_gpt} shows the height distribution of sessions collected with both policies. Random exploration fails to achieve the tallest tower, similar to experiment three from section \ref{baseline}. On the contrary, LLM-scaffolded exploration manages to find a direct path to the tallest tower in one of the sessions.

Introducing a sphere into the environment increases environmental complexity primarily because stacking cubes on top of spheres results in the cubes dropping to the table surface, resulting in a state similar to the initial state. Additionally, the presence of the sphere necessitates that it always occupies the topmost position  in the tower, adding an additional constraint. Consequently, the probability of encountering a tall tower diminishes significantly during random exploration. In contrast to the random exploration, the GPT-based approach successfully produces a tower with a height of five, revealing its ability to target hard-to-reach states.
\begin{figure}[thpb]
      \centering
      \includegraphics[width=0.48\textwidth]{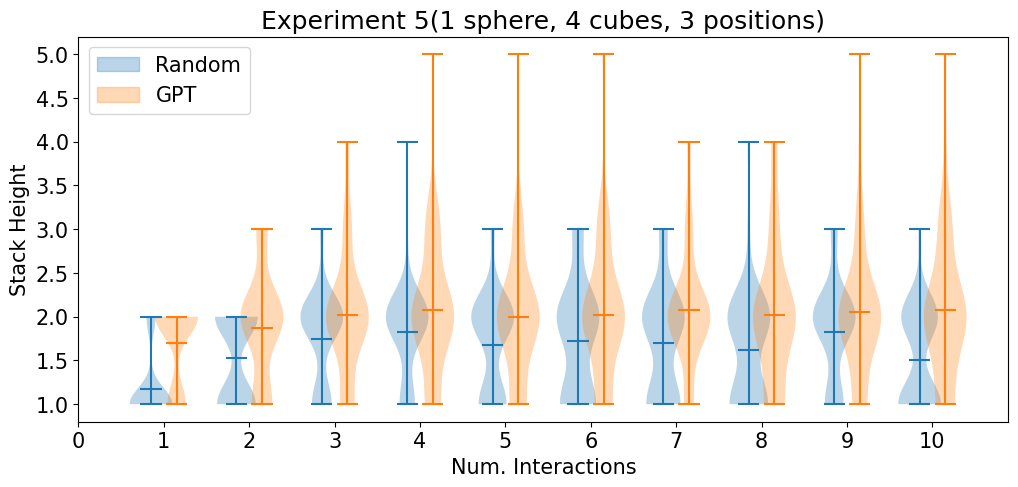}
      \caption{Height distributions of random and scaffolded sessions given 4 cubes and a sphere with 3 positions,}
      \label{sphere_rand_gpt}
\end{figure}

\begin{figure}[thpb]
        \centering
        \includegraphics[width=0.48\textwidth]{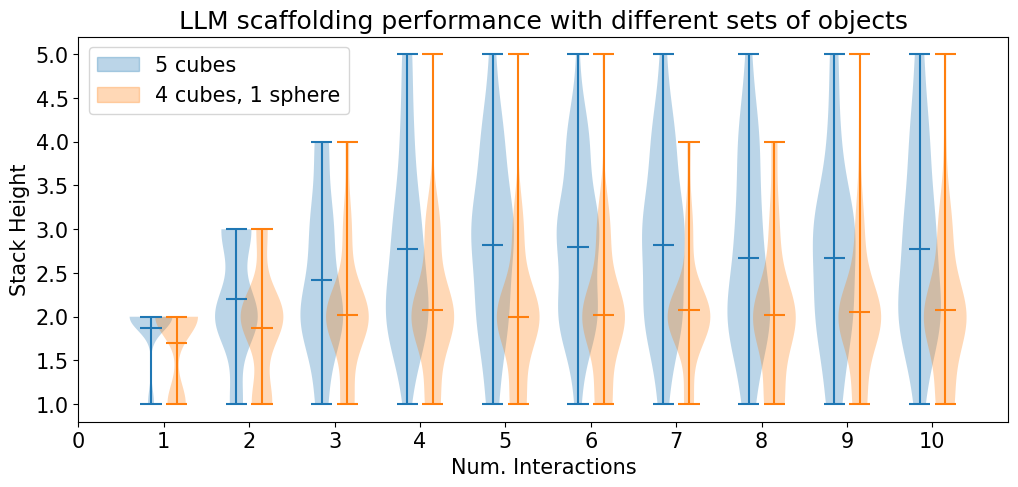}
        \caption{Comparison of average heights in LLM scaffolded sessions from experiments three and five with different sets of objects. }
        \label{gpt_vs_gpt}
   \end{figure}
Even though LLM scaffolding shows improvement over random sampling, the expectation was for the LLM to perform similarly to the experiment conducted in the environment containing five cubes by avoiding states in which the sphere is placed on an incomplete tower. However, Figure \ref{gpt_vs_gpt} compares LLM scaffolded sessions from experiments three and five and shows a decrease in average tower height when the sphere is introduced. Careful examination of dialogues during data collection reveals that GPT disregards the fact that the cubes cannot be stably stacked on the sphere and actively tries to stack cubes on top of the sphere. Moreover, it hallucinates that a cube is stacked on top of the sphere by referring to the previously executed actions section in the query prompt, even though the environment definition provides no such relation. 
\label{sec:sphere_add}
\subsection{Forcing GPT to Use Affordance}
\label{sec: tower}
During the previous experiment, GPT3.5 was conditioned to select actions based on their perceived interest. It is worth mentioning that an action leading to the construction of a tower in the long run may not be considered interesting in the current configuration. For instance, in an environment with a tower of three cubes, GPT3.5 may choose to stack one of the cubes on top of a sphere since this particular action has not been previously observed in the episode. While this behavior promotes exploration, it adds to the complexity of reaching a state with a taller tower. Furthermore, our prompt instructs GPT3.5 to select a single action, but the creation of a tower requires multi-step reasoning. 

To assess GPT3.5's understanding of affordances, we designed an experiment in which the LLM is asked to create the highest possible tower in environment configurations with 5 objects and 3 spatial positions. GPT3.5 performance in stacking task is tested in environments consisting of 5 cubes or 4 cubes and a sphere across 40 interaction sessions. During these experiments, GPT is provided with the acknowledgment that future actions will be available. For this purpose, the system prompt in the following is used while maintaining the same user prompt structure:
\\

\noindent\fbox{%
    \parbox{0.96\linewidth}{\small%
\texttt{[System]}:
There are some objects on the table. For building the highest possible tower with the objects given below, select the most appropriate manipulation action. Choose one and explain.
Your output should be in the following format:

<reasoning> some sentences </reasoning>

Selected action is : <number of the selected action>
    }
}

\begin{figure}[thpb]
      \centering
      \includegraphics[width=0.48\textwidth]{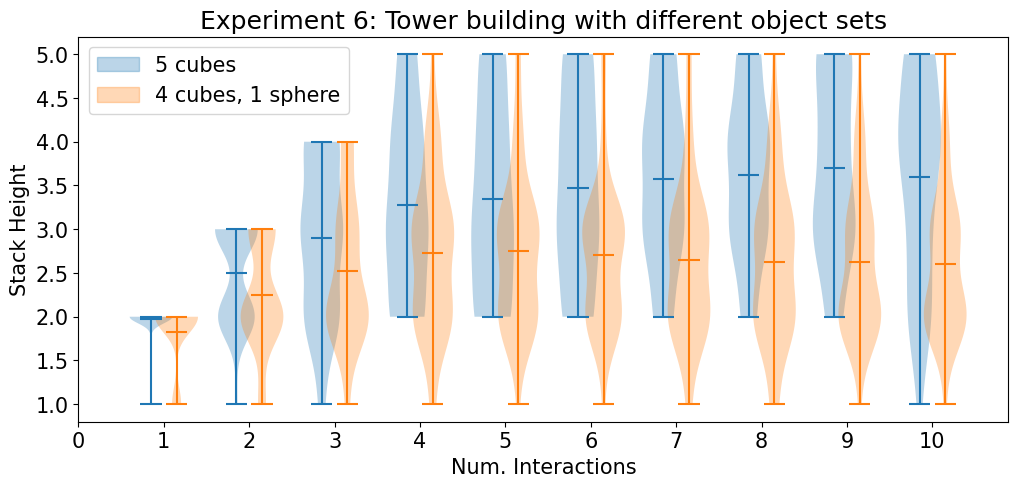}
        \caption{Comparison of average heights in different environments during LLM scaffolded sessions with tower creation task. }
        \label{pvp}
   \end{figure}
Figure \ref{pvp} illustrates the GPT's performance in these environments. Being instructed to build tall towers, the model demonstrates a preference for stacking operations, while avoiding the deconstruction of towers. As a result, higher structures are successfully built in most sessions. However, a significant difference is observed when the sphere is introduced to the environment. Further analysis reveals a continuing preference towards placing spheres in the middle positions, as this action immediately increases the height of the tower. Additionally, GPT3.5 states the sphere is a stable base for future placements to provide logical reasoning for these actions. This behavior shows that the affordances of the sphere are disregarded. Subsequently, GPT3.5  attempts to place the remaining cubes on top of the sphere, revealing a logical flaw in real world inferences. Following is an example of reasoning generated by the GPT3.5:
\\

\noindent\fbox{%
    \parbox{0.96\linewidth}{\small%
\texttt{[Assistant]}: The best action would be to put the blue cube on top of the red sphere. This is because the red sphere can provide a stable base for the cube, and the cube can sit securely on top of the sphere. 
    }
}
\\

\section{Discussion}

Compared to random exploration, LLM-scaffolded exploration shows a higher visit frequency to hard-to-reach states, even in the most complex settings in which random exploration failed to achieve the most complex structure. The clear difference between tower heights in experiments one to three and examination of the dialogue during exploration shows that GPT3.5 is capable of educated scaffolding. Additionally, the GPT actively avoids previously occurred configurations by using the session history. This is demonstrated in Experiment One and Experiment Two. In these experiments, upon creating tall structures at the beginning of the episode, GPT proceeds to dismantle them to explore proximity relations as tall towers lose their novelty, thus, a decrease in average height is observed in the later stages of the exploration. The following is an example of reasoning sampled during Experiment 2.

\vspace{1mm}
\noindent\fbox{%
    \parbox{0.96\linewidth}{\small%
\texttt{[Assistant]}: Reasoning: By putting the blue cube next to the black cube, we create a new arrangement that adds more variety to the scene. This action breaks the previous stacking pattern and introduces a new spatial relationship between the blue and black cubes.
    }
}
\vspace{2mm}

GPT3.5 can lead the exploration, however, it demonstrates a behavior change when different affordances, such as spheres, are introduced. Even though GPT3.5 is not capable of making grounded inferences with high accuracy \cite{huang2022language}, our expectation was to observe a reasonable use of affordance knowledge. When GPT3.5 is asked ``Can you balance a cube on top of a sphere?'', the generated answer indicates that it may not be feasible in the real world. However, when we tried to utilize this understanding in a tower creation task, GPT3.5 was not able to filter out episodes in which a sphere was used as a middle block in a tower. To better understand the capabilities of the GPT3.5, we asked it to generate steps to build a high tower given only the state description. The resulting steps were strongly biased toward the order of the objects appearing in the state description. Therefore, if the sphere was not the first or the last object appearing, GPT3.5 eagerly placed it in a mid position of the tower. In contrast, when GPT4 was tested in the tower creation task, it was capable of placing the sphere to the top position, regardless of the sphere's appearance order in the prompt. Moreover, some trials resulted in GPT4's discarding the sphere by stating it would not be stable even in the top position. These observations indicate a clear difference between the grounded reasoning capabilities of GPT3.5 and GPT4. This preliminary experimentation with GPT4 strongly encourages future work to investigate the exploration scaffolding capability of more powerful systems such PALM-E \cite{driess2023palm} and GPT4. Note that in this study, our goal was to see whether we can benefit robotic exploration from the harvested knowledge of GPT3.5 without {additional effort to increase grounded inference capabilities. Future work may focus on utilizing methods mentioned in section \ref{intro} alongside scaffolding.}


One limitation of our method is providing LLM with a limited set of actions. The most complex environment configuration in our experiments contains 60 possible actions. However, there are only a few \emph{correct} actions during the last stages of the tower creation task and the other actions may be neutral or adversarial. In the experiments in section \ref{sec: tower}, observation shows that the action suggestions the LLM received may be unrelated to the existing tower or the only action related to the tower may be the premature placement of the sphere in the worst case. In such cases, GPT3.5 prefers the action that immediately increases the tower height, thus, places the sphere in a mid position. Consequently, the LLM hallucinates that the sphere is a solid foundation for further stacking to validate its selection. Similarly, in the experiments from section \ref{baseline}, the average height achieved by the GPT3.5 decreases as the environment complexity increases due to the decrease in the frequency of actions resulting in a tower occurring in the prompt. One area for future work may be allowing the LLM to ask for another set of actions instead of forcing a selection. 

\section{CONCLUSIONS}
Large language models provide a valuable knowledge base for robotics applications. However, LLMs perform poorly while making grounded inferences due to their lack of real world experience. Breaching the gap between the high-level reasoning of the LLM and the low-level embodiment of the agent is essential for utilizing LLMs for robotic applications.
Different methods \cite{vemprala2023chatgpt, saycan2022arxiv, driess2023palm} are introduced to elicit grounded reasoning from LLMs. These works show remarkable success in grounded inferences, however, this success brings along an immense computation cost. To avoid these costs, LLMs can be used solely as a light knowledge base. Being inspired by parental scaffolding observed during infant development, we developed a method to utilize the LLM as a scaffolding agent. While doing so, we proposed efficient prompting strategies to tackle the LLM-embodiment alignment problem. To test the efficiency of LLM scaffolding, we drew comparisons with random exploration in different environment settings. Our experiments show the LLM's capability of actively targeting hard-to-reach states during exploration, thus, producing a robotic experience with high information gain. However, when distinct affordances are introduced to the environment, we observed an unexpected decrease in performance even though the LLM scaffolding still performs better than the random exploration.


\section*{ACKNOWLEDGMENT}
This research was supported by TUBITAK ARDEB 1001 program (project number: 120E274), Japan
Society for the Promotion of Science, Grant-in-Aid for Scientific Research – the project with number 22H03670, the International Joint
Research Promotion Program, Osaka University under the
project “Developmentally and biologically realistic modeling of perspective invariant action understanding”, and the project JPNP16007 commissioned by the New Energy and Industrial Technology Development Organization (NEDO). Additional support is provided by the University of Tokyo - IRCN Guest Researcher Program.

 \bibliographystyle{IEEEtran}
 \bibliography{refs}





\end{document}